\documentclass[letterpaper, 10 pt, conference]{ieeeconf} 

\IEEEoverridecommandlockouts                              

\newcommand{\ttitle}{Advances in Compliance Detection: Novel Models Using Vision-Based Tactile Sensors}

\usepackage{graphics} 
\usepackage{epsfig} 
\usepackage{mathptmx} 
\usepackage{times} 
\usepackage{amsmath} 
\usepackage{amssymb}  
\usepackage{multirow} 
\usepackage{xcolor}
\usepackage{makecell}
\definecolor{darkblue}{rgb}{0.00,0.00,0.70}
\usepackage[protrusion=true,expansion=true]{microtype}
\usepackage{hyperref}
\hypersetup{
  pdftitle={\ttitle},
  pdfauthor={Ziteng Li, Malte Kuhlmann, Ilana Nisky, and Nicolás Navarro-Guerrero},
  pdfsubject={Robotics (cs.RO); Artificial Intelligence (cs.AI),},%
  pdfkeywords={Tactile sensors, Young's modulus, object's compliance, Haptic and tactile perception, Machine Learning methods for robot development},%
  colorlinks=true,
  linktoc=all,
  linkcolor=darkgray,
  citecolor=darkgray,
  urlcolor=darkblue,
}

\begin{document}

\title{\ttitle}

\author{Ziteng Li$^{1*}$, Malte Kuhlmann$^{1*}$, Ilana Nisky$^{2}$, and Nicolás Navarro-Guerrero$^{1}$
\thanks{$^{1}$\href{https://www.l3s.de/}{L3S Research Center}, Leibniz Universität Hannover, Hanover, Germany
        {\tt\small \{\href{mailto:ziteng.li@l3s.de}{ziteng.li}, \href{mailto:malte.kuhlmann@l3s.de}{malte.kuhlmann}, \href{mailto:nicolas.navarro.guerrero@gmail.com}{nicolas.navarro}\}@l3s.de}}
\thanks{$^{2}$ Department of Biomedical Engineering and the School of Brain Sciences, Ben-Gurion University of the Negev, Beer Sheva, Israel {\tt\small \href{mailto:nisky@bgu.ac.il}{nisky@bgu.ac.il}}
}
\thanks{$^*$ Equal Contribution}
}

\maketitle 
\thispagestyle{empty}
\pagestyle{empty}

\begin{abstract}
Compliance is a critical parameter for describing objects in engineering, agriculture, and biomedical applications. Traditional compliance detection methods are limited by their lack of portability and scalability, rely on specialized, often expensive equipment, and are unsuitable for robotic applications. Moreover, existing neural network-based approaches using vision-based tactile sensors still suffer from insufficient prediction accuracy. In this paper, we propose two models based on Long-term Recurrent Convolutional Networks (LRCNs) and Transformer architectures that leverage RGB tactile images and other information captured by the vision-based sensor GelSight to predict compliance metrics accurately. We validate the performance of these models using multiple metrics and demonstrate their effectiveness in accurately estimating compliance. The proposed models exhibit significant performance improvement over the baseline. Additionally, we investigated the correlation between sensor compliance and object compliance estimation, which revealed that objects that are harder than the sensor are more challenging to estimate.
\end{abstract}

\section{Introduction}
In the various tasks performed by robotic dexterous hands, compliance detection for object properties is a crucial task \cite{Kappassov2015Tactile}. Compliance is a general concept related to the stiffness of an object, and it is essential for engineering design, material selection, and manufacturing processes, as it directly impacts the performance and reliability of objects in practical applications \cite{Jones2019Engineering}. It is also an important parameter for classifying objects and providing safe grasping strategies \cite{Spiers2016SingleGrasp, Toprak2018Evaluating}. In agriculture, compliance is widely used as an indicator for determining the ripeness of crops \cite{Xu2020Tactile}. Additionally, in the biomedical field, compliance is used to describe the softness of biological tissues, such as blood vessels and soft tissues \cite{Iivarinen2011Experimental}, or for designing soft robotics and tactile probes in surgical procedures for tumor detection \cite{Cianchetti2018Biomedical}. It has also been demonstrated that compliance and its perception are important in minimal invasive procedures \cite{Nisky2012Perception} and teleoperated surgeries \cite{Nisky2011Perception}.

Traditional methods for detecting object compliance often require fabricating specialized equipment or sensors tailored for specific tasks. For instance, Gao et al. \cite{Gao2024wearable} designed a piezoelectric sensor and a pneumatic actuator to measure muscle elasticity. Inoue et al.\ \cite{Inoue2020Effect} utilize a durometer to measure the object compliance of the artificial skin models used in their proposed sensor. In contrast, Britton et al.\ \cite{Britton2023Investigation} used nanoindentation testing to measure the stiffness of tissues to estimate compliance. In the area of fruit firmness, Tian et al.\ \cite{Tian2023MechanicalBased} summarize multiple types of compliance sensors based on multispectral or infrared techniques, and mechanical methods such as vibrations. Lu et al.\ \cite{Lu2006Hyperspectral} utilize hyperspectral imaging to estimate fruit firmness. However, these methods require expensive equipment and are unsuitable for robotic compliance sensing in terms of affordability and portability. 

Recently, vision-based tactile sensors have gained increasing popularity. For example, the GelSight sensor \cite{Yuan2017GelSight} is equipped with elastic-plastic surfaces that can simulate the touch variations of a human fingertip. These sensors use built-in cameras to capture RGB tactile images generated by these variations, making complex tactile information easier to analyze. Due to the low cost of these vision-based tactile sensors, they are widely used in various tasks \cite{Navarro-Guerrero2023VisuoHaptic}, including testing objects' compliance or stiffness properties. Lippi et al.\ \cite{Lippi2024LowCost} proposed a teleoperation framework for objects with varying degrees of compliance, aiming to provide tactile feedback to human operators using data from vision-based sensors installed on the robot gripper. Yuan et al.\ \cite{Yuan2017ShapeIndependent} proposed a model based on a recurrent neural network and used five images obtained from a Gelsight sensor to predict the Shore hardness as compliance information of custom-made objects. Burgess et al.\ \cite{Burgess2025Learning} utilized a gripper equipped with a Gelsight sensor to grasp objects and predicted Young's modulus of the objects by combining a convolutional neural network (CNN) with fully connected networks, along with additional information obtained from the gripper, which we also use as the baseline model. However, the existing models based on vision-based tactile sensors either require additional data beyond tactile images or their accuracy still needs improvement.

Therefore, we propose two new models for detecting object compliance based on Long-term Recurrent Convolutional Networks (LRCNs) and Transformer \cite{Kuhlmann2025VisionBased} to exploit the time-series information more effectively. Specifically, we aim to predict an object's compliance based on the feedback from the sensor when in contact with objects of varying compliance using our models. We use Young's modulus $Y$ to represent an object's compliance, as it is one of the critical indicators of compliance and a commonly used parameter in engineering design \cite{Gent1958Relation}. Additionally, in simulators like Isaac Sim \cite{Mittal2023Orbit}, Young's modulus is a key parameter for setting the compliance properties of objects. It is also an important parameter in FEM (Finite Element Method) formats, as it provides a universal definition for all materials and is typically used in well-established mechanical models \cite{deBorst2012NonLinear}.

\section{Methodology}

\subsection{Dataset}
We use the dataset released by Burgess et al.\ \cite{Burgess2025Learning} along with the Top10NN model. The model is trained to estimate Young's modulus of various real and self-made objects by grasping them with a Gelsight-equipped Franka Panda gripper. The dataset consists of the object metadata and the grasping raw data. The metadata describes the name, shape, material, and Young's modulus $\hat{Y}$ of the objects. Whereas, the raw data contains the information generated during the grasping process.

\subsubsection{Object metadata}
Object metadata includes a total of 284 objects. These objects are categorized into five different shapes and eight different materials, as shown in Fig.~\ref{fig:Dataset}. The dataset covers a wide range of materials, from foam to metal, with Young's modulus spanning nine orders of magnitude from $5 * 10^3 Pa$ to $2 * 10^{11} Pa$, representing the compliance characteristics of the majority of materials encountered in real-world robotic manipulation \cite{Huang2022DefGraspSim}.

As Young's modulus ground truth, the dataset employs two approaches. First, they used published engineering data \cite{Material} to obtain Young's modulus of the material. For objects made of materials like rubber and food, they measure the Shore hardness of the object. Then, the Shore hardness is converted to Young's modulus using Gent's hardness model \cite{Gent1958Relation} and other established methods \cite{Larson2017Can}. Meanwhile, the dataset assumes that each object is composed of a single material, so Young's modulus of the object represents the properties of that material.

\subsubsection{Raw data}
The raw data is generated from each grasp of the objects. The grasping process starts when the gripper contacts and compresses an object, stopping when the threshold of normal force of $60 N$ is reached. Each object in the dataset is grasped and recorded multiple times, which generates 7840 sets of raw data. Each raw data consists of four parts: three RGB tactile images with size, the corresponding normal force $F$, the gripper width $W$ at the current moment, and the estimates $E$ as analytical contact mode, which include $\hat{E}_{elastic}$ and $\hat{E}_{hertz}$. Specifically, the three RGB tactile images contain three frames sampled equidistantly in time, representing the frame at the start of contact, during contact, and the final frame when the threshold normal force is reached. The normal force $F$ is obtained from the force exerted on the Gelsight sensor at the current moment due to the compression of the object, while $W$ is the width of the gripper at the current moment. The analytical estimate contact mode $E$ is derived from Hooke's Law and Hertzian contact theory \cite{Fischer-Cripps1999Hertzian, Rychlewski1984Hookes}, and calculated using the normal force $F$ and width $W$, which has been used to model soft contact in robotics \cite{Dintwa2008Accuracy}, and $E$ assumes that all contacting objects are convex, simplified to a hemispherical surface. Finally, these raw data could be used as inputs for the subsequent prediction of Young's modulus.

\subsection{Experimental Design}

\begin{figure}[tbp] 
    \centering 
    \includegraphics[trim={0mm 0 0 -2mm},clip, width=\columnwidth]{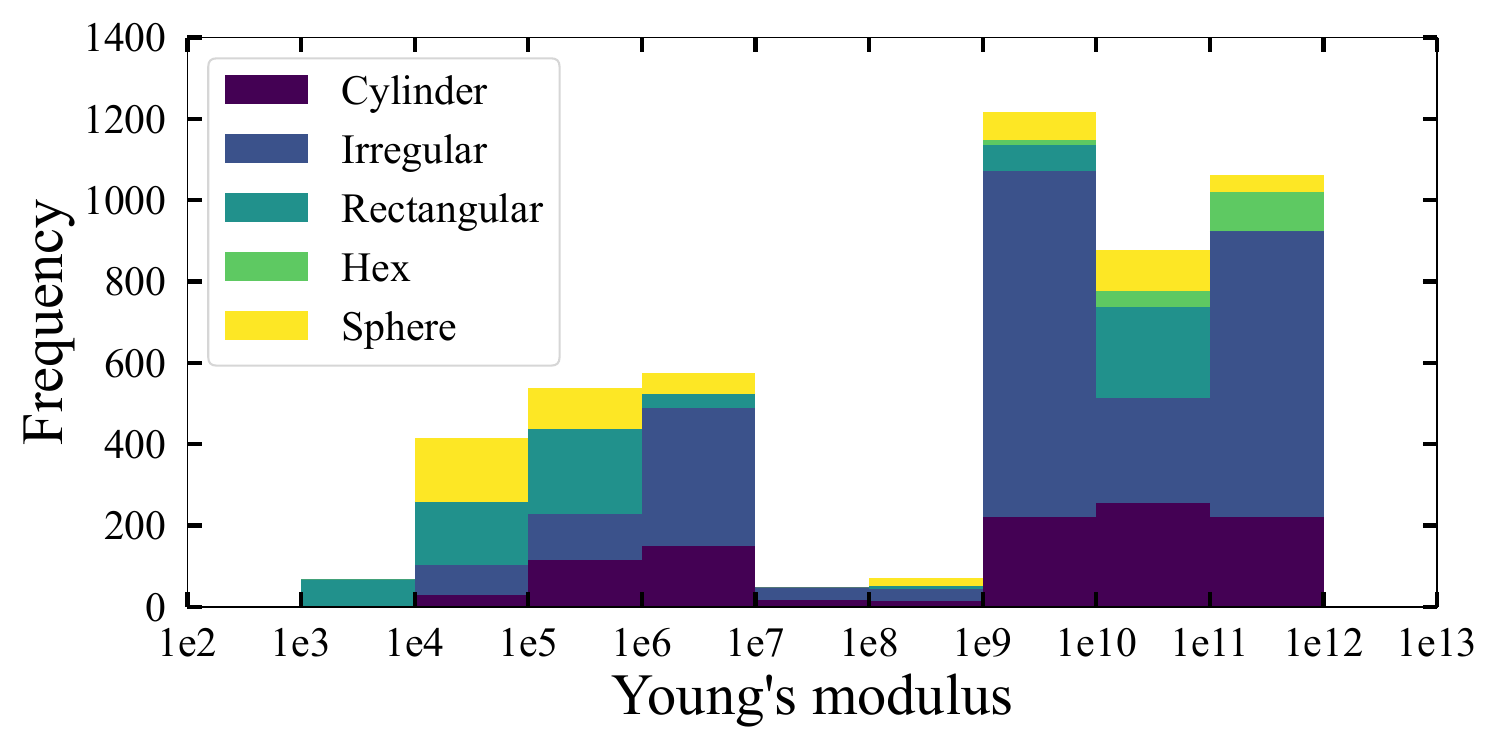}
    \caption{Dataset distribution based on normalized Young's modulus and split into object shapes.} 
    \label{fig:Dataset} 
\end{figure}

\subsubsection{Data processing}
We preprocessed the dataset by removing all samples with missing values for estimation, recorded force change, width change or objects with a single sample. The resulting dataset distribution is shown in Fig.~\ref{fig:Dataset}.

The statistics shown in Fig.~\ref{fig:Dataset} indicate that the number of objects with different shapes, materials, and Young's modulus is not uniformly distributed. We employ two strategies to sample the data and train separately. One strategy is Random sampling, which was used initially for Top10NN \cite{Burgess2025Learning}. Additionally, we introduce a balanced strategy that employs oversampling to balance the dataset according to their distribution. With these two approaches, we aim to observe the robustness of each model under different data distributions.

The dataset is divided into train, validation, and test sets. First, $20\%$ of the data is selected as the test set. The validation set comprises $20\%$ of the remaining $80\%$. The balanced strategy and data augmentation are only applied on the train and validation split. Furthermore, we apply log normalization for the original Young's modulus value, a standard method for normalizing exponential-scale data \cite{Shier2004Well}.

Moreover, we tested the models under Seen-Object and UnSeen-Object conditions. The former refers to when the objects can be present in the training, validation, and test sets. The latter indicates that the objects in the training set differ entirely from those in the validation and testing sets.

\subsubsection{Four Training strategies}
Based on the raw data, the inputs for training are divided into four components: RGB tactile images, normal force $F$, current gripper width $W$, and analytical estimates contact mode $E$. These four input modalities are generated from the Gelsight sensor, the gripper, and the analytical contact mode. 

We utilize four different input vectors to train each model separately \cite{Burgess2025Learning}: RGB tactile images only, images plus force $F$, images plus force $F$ and width $W$, and all four input modalities. We name those \textbf{Image}, \textbf{ImageF}, \textbf{ImageFW}, and \textbf{ALL} respectively. We aim to demonstrate the coupling between various inputs and Young's modulus with these combinations.

For all four training strategies, we used the Mean squared error plus $l_2$ regularization as the loss function, which is defined as Equation \ref{eq:1}. Where $y_i$ and $\hat{y_i}$ represent the log normalization values of the true Young's modulus, and the predicted Young's modulus of the object $i$, respectively.
\begin{equation}
\text{Loss} = \frac{1}{n} \sum_{i=1}^{n} (y_i - \hat{y}_i)^2 + l_2
\label{eq:1}
\end{equation}
The balanced data strategy was applied after the dataset splitting for the training and validation sets. The data was divided into nine buckets, then repeatedly resampled until a predefined threshold, $t_{balance}$, was reached. To mitigate the bias introduced by oversampling, extensive data augmentation was applied. 

We apply data augmentation consisting of random image flipping, adding Gaussian noise to the tactile image, and random color jittering. For color jittering, we set the hyperparameter to proven values used by Fu et al.\ \cite{Fu2024Touch}, and for Gaussian noise to the values used by Parag et al.\ \cite{Parag2024Learning}.

We use SMAC3 to optimize our model's hyperparameters and the balancing threshold $t_{balance}$. Due to the particularity of the task and models, we define the search space of optimized hyperparameters based on the baseline model and the general methods. We then use the optimized model to train under 10 different random seeds. All results are reported as the average of 10 training runs.

\subsection{Metrics}
\label{sec:Metrics}
Consistent with our baseline model Top10NN, we use $\log_{10}$ accuracy as one of the metrics. Specifically, the prediction is considered correct if the absolute $\log_{10}$ difference between the predicted $\hat{y}_i$ and true $y_i$ values is less or equal to 1. This metric is shown in Equation \ref{eq:2}.
\begin{equation}
\text{Prediction}_i = 
\begin{cases} 
\text{Correct}, & \text{if } \left| \log_{10}(y_i) - \log_{10}(\hat{y}_i) \right| \leq 1 \\
\text{Incorrect}, & \text{if } \left| \log_{10}(y_i) - \log_{10}(\hat{y}_i) \right| > 1
\end{cases}
\label{eq:2}
\end{equation}

This metric is designed to evaluate the model based on the order of magnitude of Young's modulus rather than the exact value. Therefore, we additionally use the Normalized Mean Squared Error (N-MSE) to evaluate the performance of all models. MSE is a standard metric for estimating a model's ability to predict continuous values in regression tasks \cite{Chicco2021Coefficient}. The motivation for using N-MSE is the wide range of Young's modulus in the dataset, spanning from $10^{3}$ to $10^{11}$. Specifically, we first applied the log transformation followed by min-max normalization to the Young's modulus in the dataset and then calculated the MSE. Thus, each model is evaluated using $\log_{10}$ accuracy and N-MSE, representing the model's performance at coarse- and fine-grained levels, respectively.

Given the variety of materials and shapes of the objects in the dataset, we also test the performance of models for each material and shape. This test aims to determine whether the model is biased towards any specific shape or material, as an ideal model should perform consistently well regardless of shape or material.

\subsection{Model Architecture}
\begin{figure}[t] 
    \centering 
    \includegraphics[trim={0mm 0 0 -6mm},clip, width=\columnwidth]{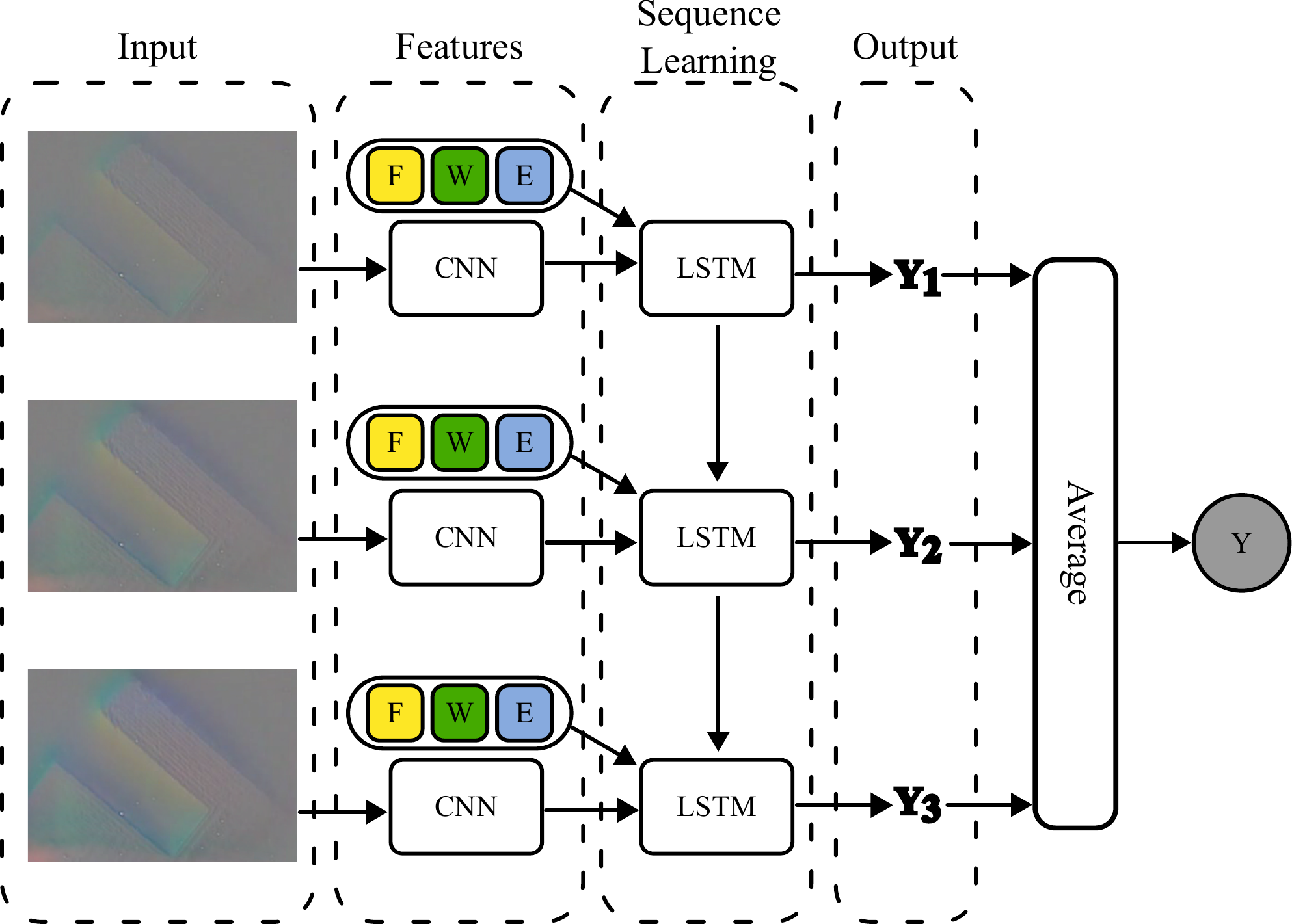} 
    \caption{VGG-LSTM Architecture.} 
    \label{fig:/LRCN} 
\end{figure}

We use Top10NN as our baseline model because it is the dataset's source. Since this model's data processing and training strategy differ from the experimental strategy proposed in the previous Subsection, we retrained it to verify its performance under our proposed experimental strategy. We also designed two new models: an LRCN-based and a Transformer-based model. The motivation for our proposed models is to exploit the time-series information in the data more effectively. We provide detailed information for each model below. The code and supplementary material are at: \url{https://github.com/mfkuhlmann/youngs-modulus-estimation}.

\begin{figure*}[t]
\centering
\includegraphics[trim={0mm 0 0 -6mm},clip, width=\textwidth]{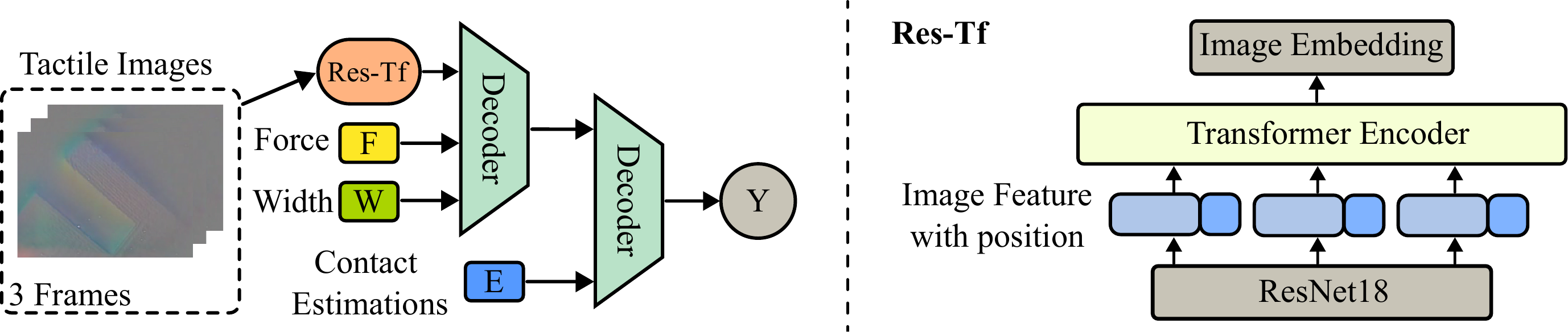} 
\caption{Res-Tf Architecture.}
\label{fig:res_tf}
\end{figure*}

\subsubsection{Top10NN}
The force $F$ and gripper width $W$ measurements are fed into Hertz and elastic analysis models to fit the estimated value $E$. Simultaneously, three frames of tactile images are input into a Convolutional Neural Network (CNN) \cite{OShea2015Introduction}. The features extracted from the images are concatenated with the grasp measurements, force $F$, and gripper width $W$, forming a large, fully connected decoder. Finally, the output features of the model are combined with the analytical estimate $E$ and sent to a smaller decoder to produce the final hybrid estimate of Young's modulus $\hat{Y}$.

\subsubsection{LRCN} 
Since the RGB tactile images belong to a set of time series data, we designed a deep learning model called \textbf{VGG-LSTM}. This model is based on Long-term Recurrent Convolutional Networks (LRCNs) \cite{Donahue2017LongTerm}. The structure of VGG-LSTM is shown in Fig.~\ref{fig:/LRCN}. Inspired by Yuan's work, our model uses CNN features from the $fc7$ layer of the VGG16 network to represent the RGB tactile images and input them into the LSTM block \cite{Yu2019Review}. Moreover, in the LSTM block, the model can optionally incorporate three additional inputs: force $F$, width $W$, and estimate $E$. Specifically, each tactile image is processed by an independent, pre-trained VGG16. The VGG16 extracts high-level features from each image, which are then fed into the corresponding LSTM block. Each LSTM block considers the current image embedding and the optional input features and integrates contextual information from the previous time step. Furthermore, each LSTM generates an output representing the features at that time step. Unlike previous research \cite{Yu2019Review}, the outputs from the three LSTMs are aggregated using learnable weighted averaging to produce the final prediction.

\subsubsection{Transformer}
We design a model named \textbf{Res-Tf} based on Residual Networks (ResNet) \cite{He2016Deep} with Transformer \cite{Vaswani2017Attention}. We incorporate a Transformer encoder due to its proven effectiveness in various time-series tasks \cite{Arnab2021ViViT}. Fig.\ \ref{fig:res_tf} shows the model's architecture. Initially, a pre-trained ResNet model is employed to extract features from each RGB tactile image, which are treated as content tokens of the image. Positional information is then integrated by concatenating each image's content token with its corresponding positional data, creating a complete token for the image, which is subsequently fed into the Transformer encoder part. An attention mechanism is applied to train the Transformer model \cite{Yuan2021TokenstoToken}. The output is combined with force, width, or analytical estimates, depending on the training strategy, and finally, the predicted Young's modulus is obtained through a decoder. 

\begin{table*}[ht]
\vspace{1.5mm}
\centering
\caption{The comparison of overall performance among all models based on Random sampling and data augmentation.}
\label{tab:random_sampling_strategy}

\renewcommand{\arraystretch}{1.15}
\begin{tabular}{|c|c|c|c|c|c|c|c|c|}

\hline
\multirow{2}{*}{\textbf{Method}} & \multicolumn{4}{c|}{\textbf{Input}} & \multicolumn{2}{c|}{\textbf{Seen-Object}} & \multicolumn{2}{c|}{\textbf{UnSeen-Object}} \\ \cline{2-9} & \centering Image & F & W & E & $\log_{10}$ accuracy& N-MSE & $\log_{10}$ accuracy& N-MSE \\ \hline \hline

\multirow{4}{*}{Top10NN} & \checkmark & - & - & -                             & 0.5199 $\pm$ 0.0258 & 0.0323 $\pm$ 0.0011 & 0.3470 $\pm$ 0.0508 & 0.0453 $\pm$ 0.0060  \\ \cline{2-9}
                          & \checkmark & \checkmark & - & -                   & 0.6042 $\pm$ 0.0153 & 0.0228 $\pm$ 0.0011 & 0.3908 $\pm$ 0.0450 & 0.0441 $\pm$ 0.0047 \\ \cline{2-9}
                          & \checkmark & \checkmark & \checkmark & -          & 0.6893 $\pm$ 0.0152 & 0.0193 $\pm$ 0.0018 & 0.5243 $\pm$ 0.0611 & 0.0339 $\pm$ 0.0047 \\ \cline{2-9}
                          & \checkmark & \checkmark & \checkmark & \checkmark & 0.5437 $\pm$ 0.0176 & 0.0286 $\pm$ 0.0011 & 0.4728 $\pm$ 0.0785 & 0.0385 $\pm$ 0.0080 \\ \Xhline{1.3px}

\multirow{4}{*}{VGG-LSTM} & \checkmark & - & - & -                            & 0.8766 $\pm$ 0.0114 & 0.0088 $\pm$ 0.0008 & 0.5695 $\pm$ 0.0334 & \textbf{0.0298} $\pm$ \textbf{0.0027} \\ \cline{2-9}
                          & \checkmark & \checkmark & - & -                   & 0.8772 $\pm$ 0.0114 & 0.0083 $\pm$ 0.0007 & 0.5715 $\pm$ 0.0546 & 0.0298 $\pm$ 0.0027 \\ \cline{2-9}
                          & \checkmark & \checkmark & \checkmark & -          & 0.8910 $\pm$ 0.0083 & 0.0079 $\pm$ 0.0010 & 0.5720 $\pm$ 0.0409 & 0.0312 $\pm$ 0.0040 \\ \cline{2-9}
                          & \checkmark & \checkmark & \checkmark & \checkmark & 0.8885 $\pm$ 0.0112 & 0.0078 $\pm$ 0.0007 & 0.5709 $\pm$ 0.0673 & 0.0308 $\pm$ 0.0039 \\ \Xhline{1.3px}

\multirow{4}{*}{Res-Tf} & \checkmark & - & - & -                              & 0.7964 $\pm$ 0.0115 & 0.0123 $\pm$ 0.0010 & 0.5844 $\pm$ 0.0422 & 0.0308 $\pm$ 0.0044 \\ \cline{2-9}
                          & \checkmark & \checkmark & - & -                   & 0.8367 $\pm$ 0.0368 & 0.0120 $\pm$ 0.0023 & 0.5900 $\pm$ 0.0453 & 0.0342 $\pm$ 0.0063 \\ \cline{2-9}
                          & \checkmark & \checkmark & \checkmark & -          & 0.8878 $\pm$ 0.0090 & 0.0077 $\pm$ 0.0007 & \textbf{0.5949} $\pm$ \textbf{0.0427} & 0.0312 $\pm$ 0.0043 \\ \cline{2-9}
                          & \checkmark & \checkmark & \checkmark & \checkmark & \textbf{0.8990} $\pm$ \textbf{0.0094} & \textbf{0.0069} $\pm$ \textbf{0.0008} & 0.5902 $\pm$ 0.0369 & 0.0321 $\pm$ 0.0045 \\ \hline

\end{tabular}
\end{table*}

\begin{table*}[ht]
\centering
\caption{The comparison of overall performance among all models based on Random sampling with the Balanced strategy and data augmentation.}
\label{tab:balanced_random_sampling_strategy}

\renewcommand{\arraystretch}{1.15}
\begin{tabular}{|c|c|c|c|c|c|c|c|c|}

\hline
\multirow{2}{*}{\textbf{Method}} & \multicolumn{4}{c|}{\textbf{Input}} & \multicolumn{2}{c|}{\textbf{Seen-Object}} & \multicolumn{2}{c|}{\textbf{UnSeen-Object}} \\ \cline{2-9} & \centering Image & F & W & E & $\log_{10}$ accuracy& N-MSE & $\log_{10}$ accuracy& N-MSE \\ \hline \hline

\multirow{4}{*}{Top10NN} & \checkmark & - & - & -                             &  $0.4583\pm0.0143$  &  $0.0346\pm0.0017$  &  $0.3274\pm0.0450$  &  $0.0475\pm0.0036$   \\ \cline{2-9}
                          & \checkmark & \checkmark & - & -                   &  $0.5832\pm0.0306$  & $0.0266\pm0.0029$  & $0.3432\pm0.0603$ & $0.0470\pm0.0046$ \\ \cline{2-9}
                          & \checkmark & \checkmark & \checkmark & -          & $0.5921\pm0.0305$ & $0.0252\pm0.0025$ & $0.4044\pm0.0596$ & $0.0381\pm0.0054$ \\ \cline{2-9}
                          & \checkmark & \checkmark & \checkmark & \checkmark & $0.4611\pm0.0145$ & $0.0315\pm0.0013$ & $0.4042\pm0.0723$ & $0.0416\pm0.0064$ \\ \Xhline{1.3px}

\multirow{4}{*}{VGG-LSTM} & \checkmark & - & - & -                            & $0.8797\pm0.0104$ & $0.0087\pm0.0008$ & $0.5550\pm0.0387$ & $\textbf{0.0301}\pm\textbf{0.0046}$ \\ \cline{2-9}
                          & \checkmark & \checkmark & - & -                   & $0.8685\pm0.0162$ & $0.0095\pm0.0009$ & $0.5295\pm0.1036$ & $0.0307\pm0.0059$ \\ \cline{2-9}
                          & \checkmark & \checkmark & \checkmark & -          & $0.8705\pm0.0131$ & $0.0088\pm0.0009$ & $0.5558\pm0.0310$ & $0.0310\pm0.0035$ \\ \cline{2-9}
                          & \checkmark & \checkmark & \checkmark & \checkmark & $\textbf{0.8822}\pm\textbf{0.0106}$ & $\textbf{0.0084}\pm\textbf{0.0010}$ & $0.5423\pm0.0774$ & $0.0302\pm0.0049$ \\ \Xhline{1.3px}

\multirow{4}{*}{Res-Tf} & \checkmark & - & - & -                              & $0.8367\pm0.0199$ & $0.0110\pm0.0016$ & $0.5191\pm0.0677$ & $0.0367\pm0.0072$ \\ \cline{2-9}
                          & \checkmark & \checkmark & - & -                   & $0.8254\pm0.0164$ & $0.0116\pm0.0014$ & $\textbf{0.5590}\pm\textbf{0.0528}$ & $0.0332\pm0.0041$ \\ \cline{2-9}
                          & \checkmark & \checkmark & \checkmark & -          & $0.8247\pm0.0181$ & $0.0112\pm0.0012$ & $0.5414\pm0.0390$ & $0.0315\pm0.0041$ \\ \cline{2-9}
                          & \checkmark & \checkmark & \checkmark & \checkmark & $0.8658\pm0.0171$ & $0.0104\pm0.0016$ & $0.5263\pm0.0800$ & $0.0323\pm0.0053$ \\ \hline

\end{tabular}
\end{table*}

\section{Results}
\label{sec:results}

\begin{figure*}[t] 
\centering
\includegraphics[trim={0mm 0 0 -6mm},clip, width=\textwidth]{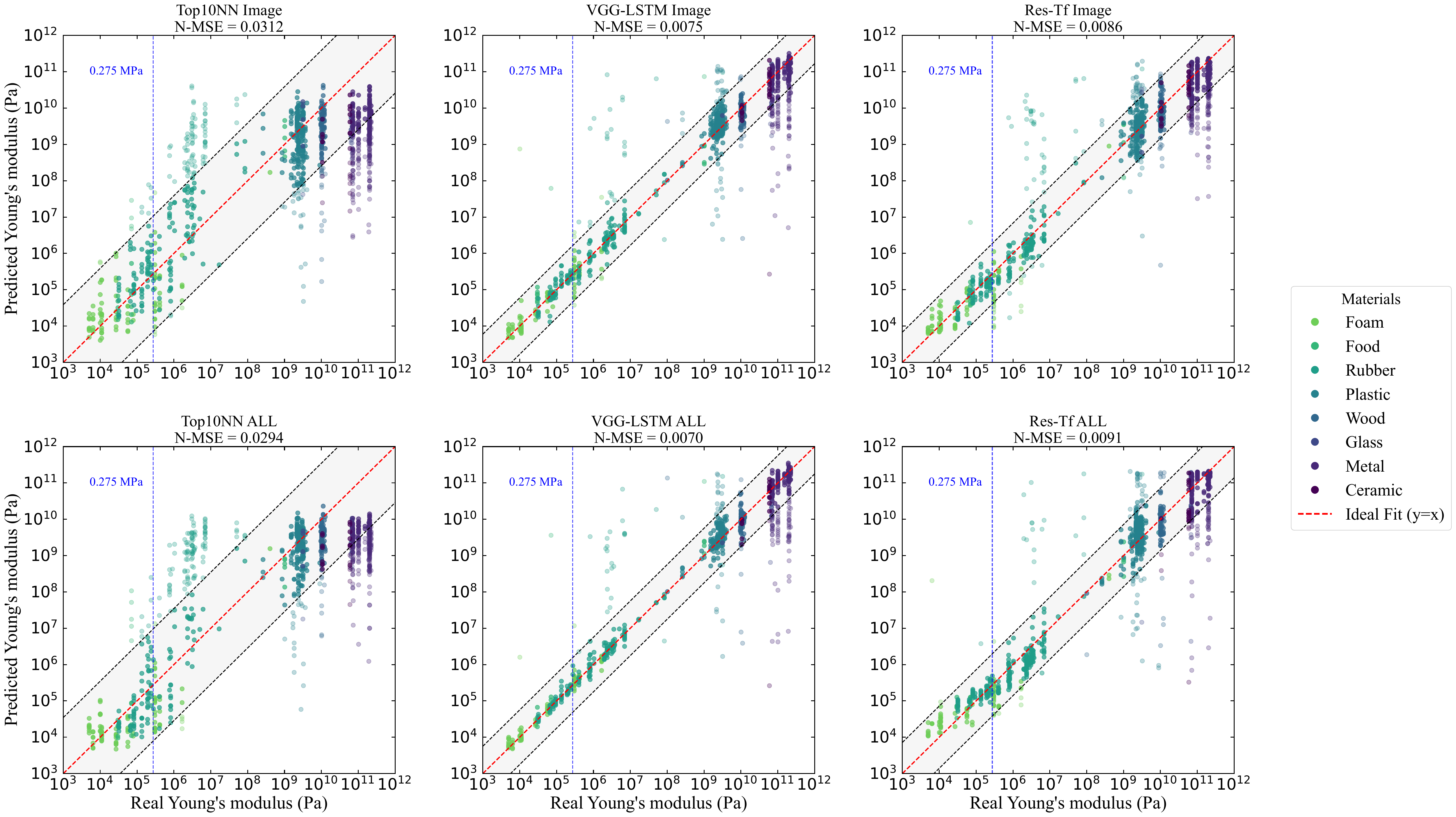} 
\caption{Comparison of Young's modulus prediction performance for the three models with two input modalities on the Seen-Object condition and Balanced strategy. The red diagonal line signifies the ground truth, while the blue vertical line denotes the Young's modulus of the GelSight sensor. The shaded area in the graph indicates the average N-MSE.}
\label{fig: Young's modulus for different materials balanced}
\end{figure*}

\subsection{Model Comparison}
\label{subsec:overall}
We first compare the $\log_{10}$ accuracy and Normalized MSE (N-MSE) performance of all models under four input training strategies: Image, ImageF, ImageFW, and ALL. We use \textbf{bold font} to denote the best results in the table.

Using the random sampling strategy, our two models outperform the baseline Top10NN in the Seen-Object and UnSeen-Object conditions, as shown in Table \ref{tab:random_sampling_strategy}. Specifically, our ALL input Res-Tf model achieved the highest $\log_{10}$ accuracy of 0.8990 and the lowest N-MSE of 0.0069 among all models on the Seen-Object condition. In the UnSeen-Object condition, our ImageFW Res-Tf model, using images, force $F$ and width $W$, achieved the highest $\log_{10}$ accuracy of 0.5949 and the VGG-LSTM with only image achieved the best N-MSE of 0.0298. In contrast, the Top10NN model does not perform competitively in either condition. Furthermore, incorporating more information for the Seen-Object condition in our two proposed models leads to a higher $\log_{10}$ accuracy and a lower N-MSE. However, in the UnSeen-Object condition, the model with less information achieves a lower N-MSE, and the differences in $\log_{10}$ accuracy between different input models are smaller in the UnSeen-Object condition compared to those in the Seen-Object condition.

The results of the experiments on the balanced dataset are shown in Table~\ref{tab:balanced_random_sampling_strategy}. The VGG-LSTM model with all inputs achieves the best performance for Seen-Object with a $\log_{10}$ accuracy of 0.8822 and an N-MSE of 0.0084. For UnSeen-Object, Res-Tf performs best in terms of $\log_{10}$ accuracy with 0.5590 and VGG-LSTM in terms of N-MSE with 0.0301. In contrast to the results presented in Table \ref{tab:random_sampling_strategy}, this analysis reveals a decline in performance for specific input combinations and an increase in performance for others. The findings indicate that the VGG-LSTM model consistently outperforms the Res-Tf model across all input combinations, suggesting that it is the more robust model for this dataset and experimental strategy.

\subsection{Material Test}
We selected six models and present their detailed performance in predicting the Young's modulus of different materials in the Seen-Object condition in Fig.\ \ref{fig: Young's modulus for different materials balanced}. For each model, we evaluated two input versions: one trained with all inputs (ALL) and another trained using only tactile images (Image), representing the maximum and minimum information usage, respectively. All six models adopted the Balanced strategy and used the seed corresponding to the lowest N-MSE. The sub-figures of Fig.\ \ref{fig: Young's modulus for different materials balanced} present the baseline Top10NN model, our VGG-LSTM model, and Res-Tf models from left to right. The first row displays models trained with only images (Image), while the second row shows models trained with all inputs (ALL). In each subfigure, each point represents a single grasp, with the x-axis and y-axis indicating the ground truth and predicted Young's modulus values, respectively. Different materials are distinguished using different colors, and the blue line represents the Young's modulus reference of  $0.275e6$ Pa for the Gelsight sensor \cite{Burgess2025Learning}. Additionally, we defined a gray-shaded region, where points within the boundary indicate that the prediction squared error (SE) for that grasp is smaller than the model's mean squared error (MSE). Each subfigure displays the N-MSE value corresponding to the respective model.

Comparing the baseline model Top10NN against our two models, VGG-LSTM and Res-Tf, we observe that the points in our two models' plots are more concentrated around the ideal fit line. The area of the margin is smaller, which means a lower N-MSE. This observation is consistent with the experimental results from the previous section. Moreover, we found that regardless of our two models used, the prediction performance is better for objects with Young's modulus lower than or close to that of the Gelsight sensor compared to objects with a higher and significantly different Young's modulus (as indicated by more points within the margin). Objects with a Young's modulus close to that of Gelsight are primarily soft materials such as foam, rubber, and food, with Young's modulus values ranging from \(10^{3}\) to \(10^{8}\). In contrast, objects with Young's modulus values that are higher and significantly different from that of Gelsight are mainly hard materials, such as wood, plastic, and metal, with values ranging from \(10^{8}\) to \(10^{12}\).

\begin{figure}[ht] 
    \centering 
    \includegraphics[width=\columnwidth]{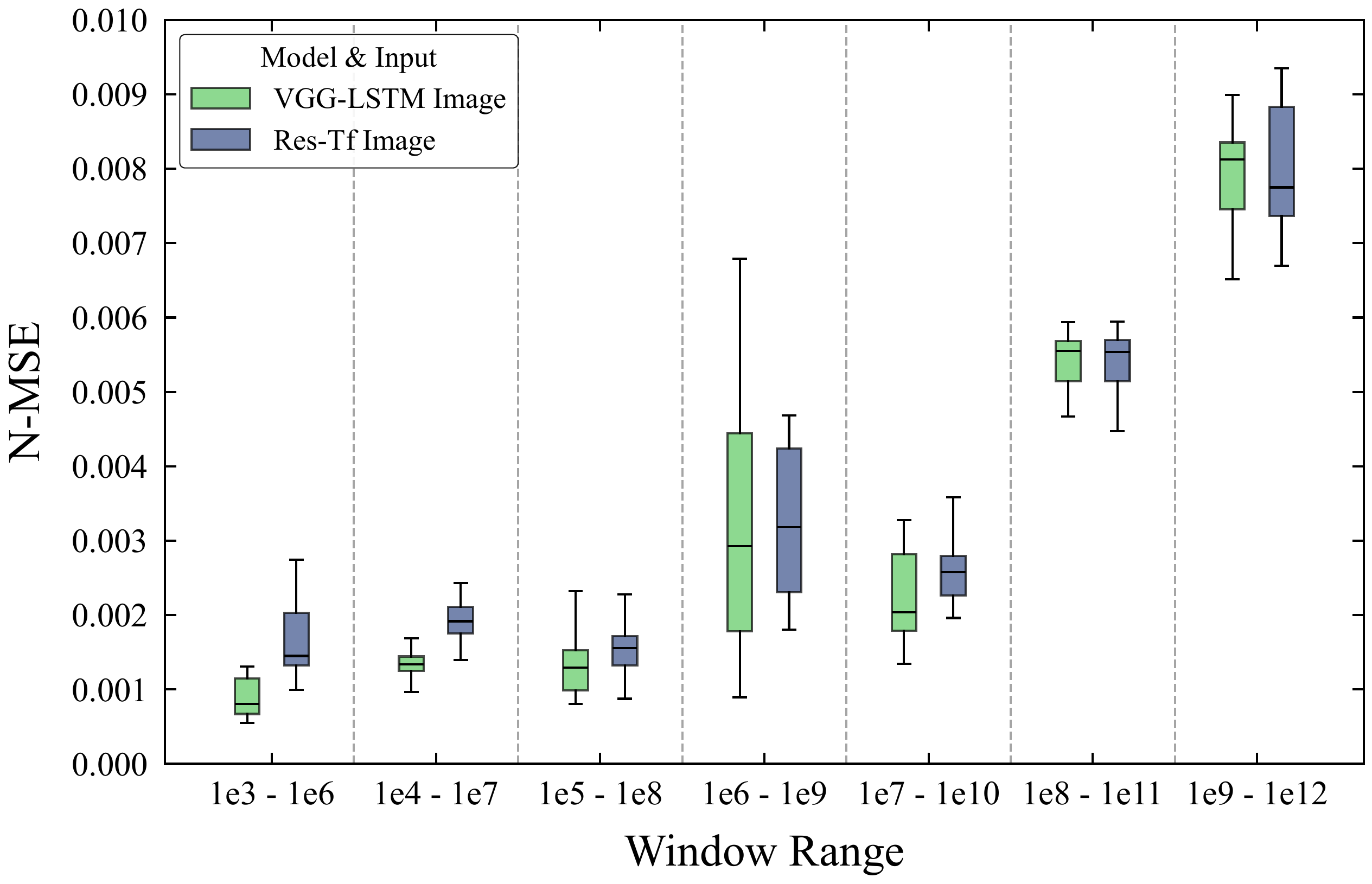} 
    \caption{N-MSE under different Young's modulus ranges based on Image only input data. Each box extends from the data's first to the third quartile, with a line at the median. The whiskers extend from each box to the farthest data point within 1.5x the interquartile range.}
    \label{fig:rolling_window} 
\end{figure}

The effect of the Young's modulus of the sensor is assessed further by evaluating the performance of two models on various material combinations. A rolling window is employed to generate seven subsets, each encompassing three magnitudes of Young's modulus. The amount of data is undersampled to ensure the same amount of data for each range. The results are presented in Fig.\ \ref{fig:rolling_window}.

Fig.\ \ref{fig:rolling_window} shows a clear trend of decreased performance with increasing magnitude. This trend begins at the threshold of $1e6 - 1e9$, where the magnitudes do not encompass data lower than the GelSight sensor. The N-MSE for each range is less than the N-MSE observed in Table \ref{tab:balanced_random_sampling_strategy}. This phenomenon occurs because the maximum error is restricted to the range of magnitude used as training data, thereby diminishing the aggregate N-MSE error. These results suggest that the accuracy of Young's modulus prediction is influenced by the hardness of the Young's modulus of the sensor.

\subsection{Shape Test}
To further verify whether the object's shape also affects the prediction of Young's modulus, we separately trained our two models on objects of different shapes and made predictions. Consistent with the previous experiments, we evaluated the performance under two input vectors: ALL and Image.

\begin{figure}[ht] 
    \centering 
    \includegraphics[width=\columnwidth]{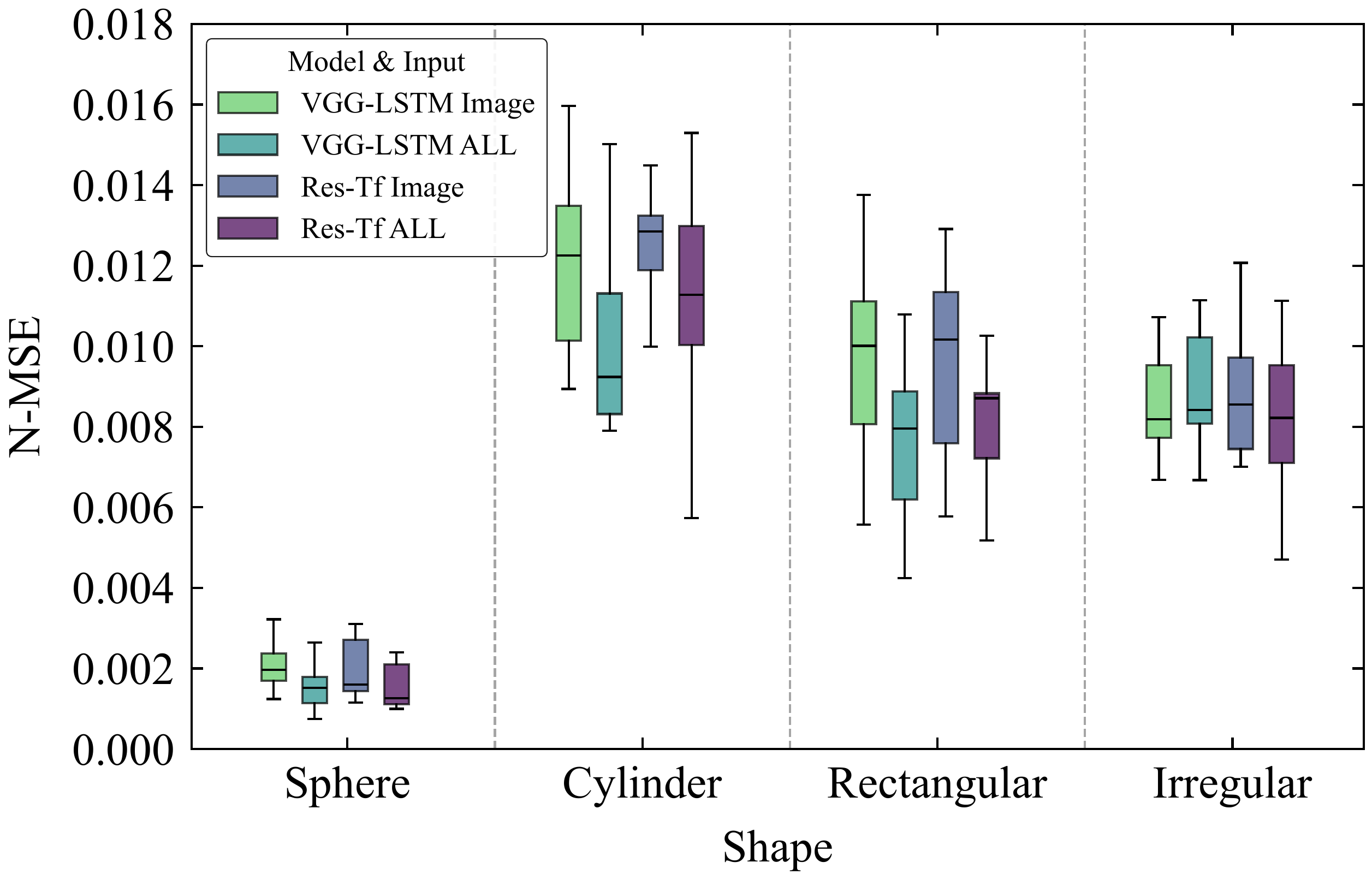}
    \caption{N-MSE under different Shapes based on Random sampling strategy with Seen-Object task. Each box extends from the data's first to the third quartile, with a line at the median. The whiskers extend from each box to the farthest data point within 1.5x the interquartile range.}
    \label{fig:shape_test} 
\end{figure}

Figure \ref{fig:shape_test} shows that all models perform best on Sphere objects and worst on Cylinder objects, consistent with the Top10NN paper. The distribution of shapes within our dataset is shown in Fig.~\ref{fig:Dataset}. Notably, Sphere and Rectangular objects tend to have softer compliance values than other shapes. Interestingly, despite having a similar distribution to Rectangular objects, Spheres exhibit better estimation performance. This result suggests that shape impacts the accuracy of compliance estimation. A reasonable explanation is that Spheres are easier to estimate due to their symmetry. Specifically, the symmetrical nature of Spheres may introduce less noise during indentation, leading to more reliable estimates.

Furthermore, our analysis reveals that the models perform worse on Cylinder objects than on Rectangular objects. This finding aligns with our previous observation that softer objects are generally easier to estimate than harder objects. Notably, the models perform similarly for Irregular and Rectangular objects despite being composed mainly of hard objects. This phenomenon could be attributed to the material composition of the Irregular objects, where certain materials are over-represented compared to others. This imbalance makes the model more susceptible to over-fitting, particularly during the Seen-Object task. In particular, the different input strategies for Irregular objects perform similarly, which is not observable for other shapes, further indicating over-fitting for Irregular shapes and the need for a more balanced dataset.

\subsection{New Dataset Test}

\begin{figure*}[t] 
\centering
\includegraphics[trim={0mm 0 0 -6mm},clip, width=\textwidth]{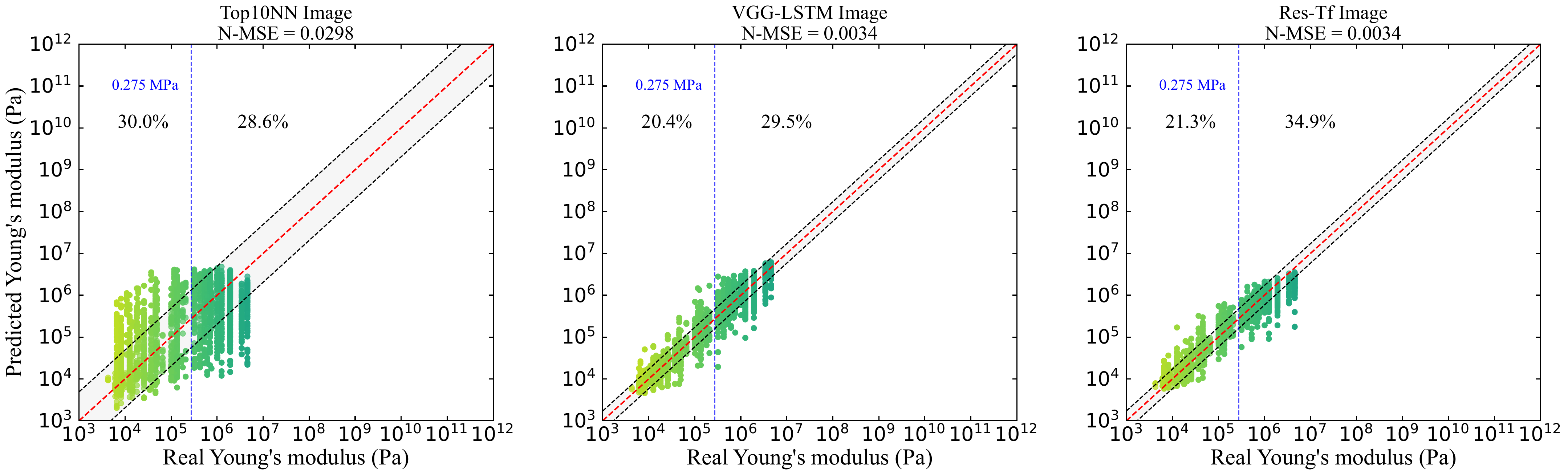} 
\caption{Comparison of Young's modulus estimation performance across all three models with Image only on Seen-Object condition for the new dataset. The red diagonal line signifies the ground truth, while the blue vertical line denotes the Young's modulus of the GelSight sensor. The shaded area in the graph indicates the average N-MSE. The percentage of predictions outside the average N-MSE is shown as floating text.}
\label{fig:new_dataset_scatter}
\end{figure*}

We conducted additional experiments using the dataset released by Yuan et al.\ \cite{Yu2019Review}. In their dataset, 371 objects of various shapes are labeled using the Shore 00 hardness scale, ranging from 8 to 87. We used an established function that converts between two measurement values to convert Shore hardness into Young's modulus to fit the input of our model \cite{Gent1958Relation} \cite{Larson2017Can}. They used video storage for each raw data sample, and we followed their method by extracting three frames as input. Since the dataset only contains raw videos, we trained our two models only to use the image as input. The data was randomly split into training, validation, and test sets. We evaluated the model using N-MSE and the $R^2$ score.

\begin{table}[!ht]
\centering
\caption{N-MSE and $R^2$ for all models trained with only Image as input on the new dataset.}
\renewcommand{\arraystretch}{1.25}
\begin{tabular}{@{}c|c c@{}}
\hline
\textbf{Model} & \textbf{N-MSE} & $\textbf{R}^\textbf{2}$\\ \hline \hline
Top10NN    & $0.0327 \pm 0.0021$ & $0.3617 \pm 0.0417$ \\ \hline
VGG-LSTM   & $0.0037 \pm 0.0002$ & $0.9275 \pm 0.0032$\\ \hline
Res-Tf     & $0.0041 \pm 0.0007$ & $0.9204 \pm 0.0131$\\ \hline
\end{tabular}
\label{tab:new_dataset_train}
\end{table}

As demonstrated in Table \ref{tab:new_dataset_train}, a similar trend to that observed in Table \ref{tab:random_sampling_strategy} can be seen, wherein our two models demonstrate significantly better performance than Top10NN. Additionally, VGG-LSTM exhibits a higher level of performance in comparison to Res-Tf. However, the performance of the models falls short of the anticipated expectations, suggesting that the new dataset poses a greater degree of difficulty than the old dataset.
A similar scatter plot as Fig.\ \ref{fig: Young's modulus for different materials balanced} was developed to provide a more comprehensive analysis. The results obtained can be found in Fig.\ \ref{fig:new_dataset_scatter}. A significant increase in bad estimations under the GelSight threshold is appreciable compared to the old dataset. Nevertheless, the proportion of bad estimations for objects below the sensor threshold is lower than for harder objects, indicating a comparable trend. This finding suggests that the observed relationship between sensor compliance and object compliance estimation is a more general issue in compliance estimation rather than a problem limited to the old dataset.

\section{Conclusions}
We propose two new models, VGG-LSTM and Res-Tf, which are based on LSTM and Transformer for predicting an object's Young's modulus. These two models can effectively leverage temporal data and achieve better results than the baseline. Additionally, we point out that regardless of the model, the tested object's relative compliance with the sensor's compliance affects prediction accuracy. Specifically, objects whose Young's modulus is closer to that of Gelsight show better performance than those whose Young's modulus is higher and distant from that of Gelsight. Furthermore, we showed that shape has some influence on the estimation performance. In addition, we highlighted the need for a more balanced and larger dataset.

Future research should prioritize creating a balanced and distributed dataset incorporating data from multiple sensors. Further investigations into the effect of different material properties, such as surface structure, on compliance estimation and the performance of other sensors, such as the DIGIT sensor, should be performed. Additionally, fine-tuning image embedding models could potentially enhance model performance. Currently, our models require an entire trajectory to estimate object compliance, which makes them impractical for real-time applications. To address this limitation, further research is needed to identify the most relevant parts of the time-series data that contain information about object compliance. This knowledge could enable a shift from requiring a whole trajectory to using only the last $n$ captured frames. Another avenue worth exploring is the viability of translating other sensor representations, such as those proposed in \cite{ZaiElAmri2024Transferring, ZaiElAmri2025DeformationBased}, to assess object stiffness.

\section*{Acknowledgment}
This research was partially funded by the Niedersächsisches Ministerium für Wissenschaft und Kultur via the Volkswagen Foundation under the Programme \href{https://zukunft.niedersachsen.de/foerderangebot/forschungskooperation-niedersachsen-israel/}{zukunft.niedersachsen: Forschungskooperation Niedersachsen -- Israel} project No.\ 15-76251-5616/2023 (\href{https://nicolas-navarro-guerrero.github.io/projects/romeo/}{ROMEO}).

\bibliographystyle{IEEEtran}
\bibliography{paper}

\begin{thebibliography}{10}
\providecommand{\url}[1]{#1}
\csname url@samestyle\endcsname
\providecommand{\newblock}{\relax}
\providecommand{\bibinfo}[2]{#2}
\providecommand{\BIBentrySTDinterwordspacing}{\spaceskip=0pt\relax}
\providecommand{\BIBentryALTinterwordstretchfactor}{4}
\providecommand{\BIBentryALTinterwordspacing}{\spaceskip=\fontdimen2\font plus
\BIBentryALTinterwordstretchfactor\fontdimen3\font minus
  \fontdimen4\font\relax}
\providecommand{\BIBforeignlanguage}[2]{{%
\expandafter\ifx\csname l@#1\endcsname\relax
\typeout{** WARNING: IEEEtran.bst: No hyphenation pattern has been}%
\typeout{** loaded for the language `#1'. Using the pattern for}%
\typeout{** the default language instead.}%
\else
\language=\csname l@#1\endcsname
\fi
#2}}
\providecommand{\BIBdecl}{\relax}
\BIBdecl

\bibitem{Kappassov2015Tactile}
Z.~Kappassov, J.-A. Corrales, and V.~Perdereau, ``Tactile {{Sensing}} in
  {{Dexterous Robot Hands}} -- {{Review}},'' \emph{Robotics and Autonomous
  Systems}, vol. 74, Part A, pp. 195--220, Dec. 2015.

\bibitem{Jones2019Engineering}
D.~R.~H. Jones and M.~F. Ashby, \emph{Engineering {{Materials}} 1: {{An
  Introduction}} to {{Properties}}, {{Applications}} and {{Design}}},
  5th~ed.\hskip 1em plus 0.5em minus 0.4em\relax Oxford, United Kingdom:
  Butterworth-Heinemann, 2019, vol.~1.

\bibitem{Spiers2016SingleGrasp}
A.~J. Spiers, M.~V. Liarokapis, B.~Calli, and A.~M. Dollar, ``Single-{{Grasp
  Object Classification}} and {{Feature Extraction}} with {{Simple Robot
  Hands}} and {{Tactile Sensors}},'' \emph{IEEE Transactions on Haptics},
  vol.~9, no.~2, pp. 207--220, 2016.

\bibitem{Toprak2018Evaluating}
S.~Toprak, N.~{Navarro-Guerrero}, and S.~Wermter, ``Evaluating {{Integration
  Strategies}} for {{Visuo-Haptic Object Recognition}},'' \emph{Cognitive
  Computation}, vol.~10, no.~3, pp. 408--425, Jun. 2018.

\bibitem{Xu2020Tactile}
C.~Xu, H.~He, S.~C. Hauser, and G.~J. Gerling, ``Tactile {{Exploration
  Strategies With Natural Compliant Objects Elicit Virtual Stiffness Cues}},''
  \emph{IEEE Transactions on Haptics}, vol.~13, no.~1, pp. 4--10, Jan. 2020.

\bibitem{Iivarinen2011Experimental}
J.~T. Iivarinen, R.~K. Korhonen, P.~Julkunen, and J.~S. Jurvelin,
  ``Experimental and {{Computational Analysis}} of {{Soft Tissue Stiffness}} in
  {{Forearm Using}} a {{Manual Indentation Device}},'' \emph{Medical
  Engineering \& Physics}, vol.~33, no.~10, pp. 1245--1253, Dec. 2011.

\bibitem{Cianchetti2018Biomedical}
M.~Cianchetti, C.~Laschi, A.~Menciassi, and P.~Dario, ``Biomedical
  {{Applications}} of {{Soft Robotics}},'' \emph{Nature Reviews Materials},
  vol.~3, no.~6, pp. 143--153, Jun. 2018.

\bibitem{Nisky2012Perception}
I.~Nisky, F.~Huang, A.~Milstein, C.~M. Pugh, F.~A. {Mussa-ivaldi}, and
  A.~Karniel, ``Perception of {{Stiffness}} in {{Laparoscopy}} -- the {{Fulcrum
  Effect}},'' \emph{Studies in health technology and informatics}, vol. 173,
  pp. 313--319, 2012.

\bibitem{Nisky2011Perception}
I.~Nisky, A.~Pressman, C.~M. Pugh, F.~A. {Mussa-Ivaldi}, and A.~Karniel,
  ``Perception and {{Action}} in {{Teleoperated Needle Insertion}},''
  \emph{IEEE Transactions on Haptics}, vol.~4, no.~3, pp. 155--166, Jul. 2011.

\bibitem{Gao2024wearable}
D.~Gao, J.~P. Lee, J.~Chen, L.~S. Tay, Y.~Xin, K.~Parida, M.~W.~M. Tan,
  P.~Huang, K.~H. Kong, and P.~S. Lee, ``A {{Wearable Pneumatic-Piezoelectric
  System}} for {{Quantitative Assessment}} of {{Skeletomuscular
  Biomechanics}},'' \emph{Device}, vol.~2, no.~3, p. 100288, Mar. 2024.

\bibitem{Inoue2020Effect}
K.~Inoue, S.~Okamoto, Y.~Akiyama, and Y.~Yamada, ``Effect of {{Material
  Hardness}} on {{Friction Between}} a {{Bare Finger}} and {{Dry}} and
  {{Lubricated Artificial Skin}},'' \emph{IEEE Transactions on Haptics},
  vol.~13, no.~1, pp. 123--129, Jan. 2020.

\bibitem{Britton2023Investigation}
M.~Britton, E.~Parle, and T.~J. Vaughan, ``An {{Investigation}} on the
  {{Effects}} of in {{Vitro Induced Advanced Glycation End-Products}} on
  {{Cortical Bone Fracture Mechanics}} at {{Fall-Related Loading Rates}},''
  \emph{Journal of the Mechanical Behavior of Biomedical Materials}, vol. 138,
  p. 105619, Feb. 2023.

\bibitem{Tian2023MechanicalBased}
S.~Tian and H.~Xu, ``Mechanical-{{Based}} and {{Optical-Based Methods}} for
  {{Nondestructive Evaluation}} of {{Fruit Firmness}},'' \emph{Food Reviews
  International}, vol.~39, no.~7, pp. 4009--4039, Aug. 2023.

\bibitem{Lu2006Hyperspectral}
R.~Lu and Y.~Peng, ``Hyperspectral {{Scattering}} for {{Assessing Peach Fruit
  Firmness}},'' \emph{Biosystems Engineering}, vol.~93, no.~2, pp. 161--171,
  Feb. 2006.

\bibitem{Yuan2017GelSight}
W.~Yuan, S.~Dong, and E.~H. Adelson, ``{{GelSight}}: {{High-Resolution Robot
  Tactile Sensors}} for {{Estimating Geometry}} and {{Force}},''
  \emph{Sensors}, vol.~17, no.~12, p. 2762, Dec. 2017.

\bibitem{Navarro-Guerrero2023VisuoHaptic}
N.~{Navarro-Guerrero}, S.~Toprak, J.~Josifovski, and L.~Jamone,
  ``Visuo-{{Haptic Object Perception}} for {{Robots}}: {{An Overview}},''
  \emph{Autonomous Robots}, vol.~47, no.~4, pp. 377--403, Apr. 2023.

\bibitem{Lippi2024LowCost}
M.~Lippi, M.~C. Welle, M.~K. Wozniak, A.~Gasparri, and D.~Kragic, ``Low-{{Cost
  Teleoperation}} with {{Haptic Feedback}} through {{Vision-based Tactile
  Sensors}} for {{Rigid}} and {{Soft Object Manipulation}},'' arXiv, Tech. Rep.
  arXiv:2403.16764, Mar. 2024.

\bibitem{Yuan2017ShapeIndependent}
W.~Yuan, C.~Zhu, A.~Owens, M.~A. Srinivasan, and E.~H. Adelson,
  ``Shape-{{Independent Hardness Estimation Using Deep Learning}} and a
  {{Gelsight Tactile Sensor}},'' in \emph{{{IEEE International Conference}} on
  {{Robotics}} and {{Automation}} ({{ICRA}})}.\hskip 1em plus 0.5em minus
  0.4em\relax Singapore: IEEE, May 2017, pp. 951--958.

\bibitem{Burgess2025Learning}
M.~Burgess, J.~Zhao, and L.~Willemet, ``Learning {{Object Compliance}} via
  {{Young}}'s {{Modulus}} from {{Single Grasps}} using {{Camera-Based Tactile
  Sensors}},'' arXiv, Tech. Rep. arXiv:2406.15304, 2025.

\bibitem{Kuhlmann2025VisionBased}
M.~Kuhlmann, Z.~Li, and N.~{Navarro-Guerrero}, ``Toward {{Vision-Based Object
  Compliance Estimation}},'' in \emph{German {{Robotics Conference}}
  ({{GRC}})}, ser. 1st, Nuremberg, Germany, Mar. 2025, pp. 1--3.

\bibitem{Gent1958Relation}
A.~N. Gent, ``On the {{Relation}} between {{Indentation Hardness}} and
  {{Young}}'s {{Modulus}},'' \emph{Rubber Chemistry and Technology}, vol.~31,
  no.~4, pp. 896--906, Sep. 1958.

\bibitem{Mittal2023Orbit}
M.~Mittal, C.~Yu, Q.~Yu, J.~Liu, N.~Rudin, D.~Hoeller, J.~L. Yuan, R.~Singh,
  Y.~Guo, H.~Mazhar, A.~Mandlekar, B.~Babich, G.~State, M.~Hutter, and A.~Garg,
  ``Orbit: {{A Unified Simulation Framework}} for {{Interactive Robot Learning
  Environments}},'' \emph{IEEE Robotics and Automation Letters}, vol.~8, no.~6,
  pp. 3740--3747, Jun. 2023.

\bibitem{deBorst2012NonLinear}
R.~de~Borst, M.~A. Crisfield, J.~J.~C. Remmers, and C.~V. Verhoosel,
  \emph{Non-{{Linear Finite Element Analysis}} of {{Solids}} and
  {{Structures}}}, 2nd~ed., ser. Wiley Series in Computational Mechanics.\hskip
  1em plus 0.5em minus 0.4em\relax Chichester, West Sussex, United Kingdom:
  Wiley, 2012.

\bibitem{Huang2022DefGraspSim}
I.~Huang, Y.~Narang, C.~Eppner, B.~Sundaralingam, M.~Macklin, R.~Bajcsy,
  T.~Hermans, and D.~Fox, ``{{DefGraspSim}}: {{Physics-Based Simulation}} of
  {{Grasp Outcomes}} for {{3D Deformable Objects}},'' \emph{IEEE Robotics and
  Automation Letters}, vol.~7, no.~3, pp. 6274--6281, 2022.

\bibitem{Material}
\BIBentryALTinterwordspacing
MatWeb, ``Material property data,'' 2024. [Online]. Available:
  \url{https://www.matweb.com}
\BIBentrySTDinterwordspacing

\bibitem{Larson2017Can}
K.~Larson, ``Can {{You Estimate Modulus}} from {{Durometer Hardness}} for
  {{Silicones}}? {{Yes}}, but {{Only Roughly}} {\dots} and {{You Must Choose
  Your Modulus Carefully}}!'' Dow Chemical Company, White Paper, 2017.

\bibitem{Fischer-Cripps1999Hertzian}
A.~C. {Fischer-Cripps}, ``The {{Hertzian Contact Surface}},'' \emph{Journal of
  Materials Science}, vol.~34, no.~1, pp. 129--137, Jan. 1999.

\bibitem{Rychlewski1984Hookes}
J.~Rychlewski, ``On {{Hooke}}'s law,'' \emph{Journal of Applied Mathematics and
  Mechanics}, vol.~48, no.~3, pp. 303--314, Jan. 1984.

\bibitem{Dintwa2008Accuracy}
E.~Dintwa, E.~Tijskens, and H.~Ramon, ``On the {{Accuracy}} of the {{Hertz
  Model}} to {{Describe}} the {{Normal Contact}} of {{Soft Elastic Spheres}},''
  \emph{Granular Matter}, vol.~10, no.~3, pp. 209--221, Mar. 2008.

\bibitem{Shier2004Well}
D.~E. Shier, ``Well {{Log Normalization}}: {{Methods}} and {{Guidelines}},''
  \emph{Petrophysics - The SPWLA Journal}, vol.~45, no.~03, May 2004.

\bibitem{Fu2024Touch}
L.~Fu, G.~Datta, H.~Huang, W.~C.-H. Panitch, J.~Drake, J.~Ortiz, M.~Mukadam,
  M.~Lambeta, R.~Calandra, and K.~Goldberg, ``A {{Touch}}, {{Vision}}, and
  {{Language Dataset}} for {{Multimodal Alignment}},'' in \emph{International
  {{Conference}} on {{Machine Learning}} ({{ICML}})}, ser. {{ICML}}'24, vol.
  235.\hskip 1em plus 0.5em minus 0.4em\relax Vienna, Austria: JMLR.org, Jul.
  2024, pp. 14\,080--14\,101.

\bibitem{Parag2024Learning}
A.~Parag, E.~H. Adelson, and E.~Misimi, ``Learning {{Incipient Slip}} with
  {{Gelsight Sensors}}: {{Attention Classification}} with {{Video Vision
  Transformers}},'' in \emph{{{IEEE}}/{{RSJ International Conference}} on
  {{Intelligent Robots}} and {{Systems}} ({{IROS}})}, Oct. 2024, pp.
  13\,960--13\,966.

\bibitem{Chicco2021Coefficient}
D.~Chicco, M.~J. Warrens, and G.~Jurman, ``The {{Coefficient}} of
  {{Determination R-Squared Is More Informative Than SMAPE}}, {{MAE}},
  {{MAPE}}, {{MSE}} and {{RMSE}} in {{Regression Analysis Evaluation}},''
  \emph{PeerJ Computer Science}, vol.~7, p. e623, Jul. 2021.

\bibitem{OShea2015Introduction}
K.~O'Shea and R.~Nash, ``An {{Introduction}} to {{Convolutional Neural
  Networks}},'' arXiv, Tech. Rep. arXiv:1511.08458, Dec. 2015.

\bibitem{Donahue2017LongTerm}
J.~Donahue, L.~A. Hendricks, M.~Rohrbach, S.~Venugopalan, S.~Guadarrama,
  K.~Saenko, and T.~Darrell, ``Long-{{Term Recurrent Convolutional Networks}}
  for {{Visual Recognition}} and {{Description}},'' \emph{IEEE Transactions on
  Pattern Analysis and Machine Intelligence}, vol.~39, no.~04, pp. 677--691,
  Apr. 2017.

\bibitem{Yu2019Review}
Y.~Yu, X.~Si, C.~Hu, and J.~Zhang, ``A {{Review}} of {{Recurrent Neural
  Networks}}: {{LSTM Cells}} and {{Network Architectures}},'' \emph{Neural
  Computation}, vol.~31, no.~7, pp. 1235--1270, Jul. 2019.

\bibitem{He2016Deep}
K.~He, X.~Zhang, S.~Ren, and J.~Sun, ``Deep {{Residual Learning}} for {{Image
  Recognition}},'' in \emph{{{IEEE Conference}} on {{Computer Vision}} and
  {{Pattern Recognition}} ({{CVPR}})}, Las Vegas, NV, USA, 2016, pp. 770--778.

\bibitem{Vaswani2017Attention}
A.~Vaswani, N.~Shazeer, N.~Parmar, J.~Uszkoreit, L.~Jones, A.~N. Gomez,
  {\L}.~Kaiser, and I.~Polosukhin, ``Attention {{Is All You Need}},'' in
  \emph{Advances in {{Neural Information Processing Systems}} ({{NIPS}})},
  vol.~30, Long Beach, CA, USA, 2017, p.~11.

\bibitem{Arnab2021ViViT}
A.~Arnab, M.~Dehghani, G.~Heigold, C.~Sun, M.~Lu{\v c}i{\'c}, and C.~Schmid,
  ``{{ViViT}}: {{A Video Vision Transformer}},'' in \emph{{{IEEE}}/{{CVF
  International Conference}} on {{Computer Vision}} ({{ICCV}})}, Oct. 2021, pp.
  6816--6826.

\bibitem{Yuan2021TokenstoToken}
L.~Yuan, Y.~Chen, T.~Wang, W.~Yu, Y.~Shi, Z.~Jiang, F.~E.~H. Tay, J.~Feng, and
  S.~Yan, ``Tokens-to-{{Token ViT}}: {{Training Vision Transformers}} from
  {{Scratch}} on {{ImageNet}},'' in \emph{{{IEEE}}/{{CVF International
  Conference}} on {{Computer Vision}} ({{ICCV}})}, Montreal, QC, Canada, 2021,
  pp. 538--547.

\bibitem{ZaiElAmri2024Transferring}
W.~Zai El~Amri, M.~Kuhlmann, and N.~{Navarro-Guerrero}, ``Transferring
  {{Tactile Data Across Sensors}},'' in \emph{40th {{Anniversary}} of the
  {{IEEE Conference}} on {{Robotics}} and {{Automation}} ({{ICRA}}@40)},
  Rotterdam, The Netherlands, Sep. 2024, pp. 1540--1542.

\bibitem{ZaiElAmri2025DeformationBased}
------, ``{{ACROSS}}: {{A Deformation-Based Cross-Modal Representation}} for
  {{Robotic Tactile Perception}},'' in \emph{{{IEEE International Conference}}
  on {{Robotics}} and {{Automation}} ({{ICRA}})}, Atlanta, GA, USA, 2025, pp.
  1--8.

\end{thebibliography}

\end{document}